\documentclass[conference]{IEEEtran}
\IEEEoverridecommandlockouts

\usepackage{multirow}
\usepackage{makecell}
\usepackage[table,xcdraw]{xcolor}
\usepackage{array}
\usepackage{graphicx,psfrag}
\usepackage{caption}
\usepackage{tikz}
\usepackage{subfig} 
\usepackage{pifont}
\usepackage{relsize}
\usepackage{soul,color}
\soulregister\cite7
\soulregister\ref7
\soulregister\pageref7
\usepackage{amssymb}
\usepackage{balance}
\usepackage{amsmath}
\usepackage{url}
\usepackage{algpseudocode}
\usepackage{booktabs}
\renewcommand\arraystretch{1.6}
\ifCLASSOPTIONcompsoc
\usepackage[nocompress]{cite}
\else
\usepackage{cite}
\fi
\ifCLASSINFOpdf
\else
\fi

\ifCLASSINFOpdf
\else

\fi

\begin{document}

\title{Integration of Domain Knowledge using Medical Knowledge Graph Deep Learning \\for Cancer Phenotyping}


\author{\IEEEauthorblockN{Mohammed Alawad\IEEEauthorrefmark{1},
Shang Gao\IEEEauthorrefmark{2},
Mayanka Chandra Shekar\IEEEauthorrefmark{2},
S.M.Shamimul Hasan\IEEEauthorrefmark{2},
J. Blair Christian\IEEEauthorrefmark{2},
Xiao-Cheng Wu\IEEEauthorrefmark{3},\\
Eric B. Durbin\IEEEauthorrefmark{4},
Jennifer Doherty\IEEEauthorrefmark{5},
Antoinette Stroup\IEEEauthorrefmark{6},
Linda Coyle\IEEEauthorrefmark{7},
Lynne Penberthy\IEEEauthorrefmark{8},
Georgia Tourassi\IEEEauthorrefmark{9}}\\
\IEEEauthorblockA{\IEEEauthorrefmark{1}Electrical and Computer Engineering Department, Wayne State University, Detroit, MI, USA.}
\IEEEauthorblockA{\IEEEauthorrefmark{2}Computational Sciences and Engineering Division, Oak Ridge National Laboratory, Oak Ridge, TN, USA.}
\IEEEauthorblockA{\IEEEauthorrefmark{3}Louisiana Tumor Registry, Louisiana State University Health Sciences Center, New Orleans, LA, USA.}
\IEEEauthorblockA{\IEEEauthorrefmark{4}Kentucky Cancer Registry and College of Medicine, University of Kentucky, Lexington, KY, USA.}
\IEEEauthorblockA{\IEEEauthorrefmark{5}Utah Cancer Registry, University of Utah School of Medicine, Salt Lake City, Utah, USA}
\IEEEauthorblockA{\IEEEauthorrefmark{6}New Jersey State Cancer Registry, New Jersey Department of Health, Trenton, New Jersey, USA}
\IEEEauthorblockA{\IEEEauthorrefmark{7}Information Management Services Inc, Calverton, MD, USA.}
\IEEEauthorblockA{\IEEEauthorrefmark{8}Surveillance Research Program, Division of Cancer Control and Population Sciences,\\ National Cancer Institute, Bethesda, MD, USA.}
\IEEEauthorblockA{\IEEEauthorrefmark{9} National Center for Computational Sciences, Oak Ridge National Laboratory, Oak Ridge, TN, USA.}
Email: \{alawad\}@wayne.edu

\thanks{This manuscript has been authored by UT
- Battelle, LLC under Contract No. DE-AC05-00OR22725 with the U.S. Department of Energy. The United States 
Government retains and the publisher, by accepting the article for publication, acknowledges that the United States Government retains a non-exclusive, paid-
up, irrevocable, world-wide license to publish or reproduce the published form of the manuscript, or allow others to do so, for United States Government purposes. The Department of Energy will provide public access to these results of federally sponsored research in accordance with the DOE Public Access Plan (http://energy.gov/downloads/doe-public-
access-plan).}
}

\maketitle

\begin{abstract}
A key component of deep learning (DL) for natural language processing (NLP) is word embeddings. Word embeddings that effectively capture the meaning and context of the word that they represent can significantly improve the performance of downstream DL models for various NLP tasks. Many existing word embeddings techniques capture the context of words based on word co-occurrence in documents and text; however, they often cannot capture broader domain-specific relationships between concepts that may be crucial for the NLP task at hand. In this paper, we propose a method to integrate external knowledge from medical terminology ontologies into the context captured by word embeddings. Specifically, we use a medical knowledge graph, such as the unified medical language system (UMLS), to find connections between clinical terms in cancer pathology reports. This approach aims to minimize the distance between connected clinical concepts. We evaluate the proposed approach using a Multitask Convolutional Neural Network (MT-CNN) to extract six cancer characteristics -- site, subsite, laterality, behavior, histology, and grade -- from a dataset of ~900K cancer pathology reports. The results show that the MT-CNN model which uses our domain informed embeddings outperforms the same MT-CNN using standard word2vec embeddings across all tasks, with an improvement in the overall micro- and macro-F1 scores by 4.97\%and 22.5\%, respectively.

\end{abstract}


\IEEEpeerreviewmaketitle

\section{Introduction}
\label{section:intro}

Distributed word representations have become an essential step towards the development of any automated information extraction tool. The de facto word embedding techniques are generated based on the word level information present in a large corpus of text data within a given domain. Word embedding techniques have been effective in capturing the semantic information through observed similarities in context it was trained on, but it lacks the relational information between clinical concepts, which is essential for the biomedical domain.

For creating clinically relevant and useful embeddings, there has been an increase in the use of transfer learning \cite{howard2018universal}. Word embeddings are first trained using a vast corpus of unlabeled text data, and these pre-trained embeddings are used as inputs for a downstream supervised task. In the context of cancer phenotype classification, we first train word embeddings using a large corpus of cancer pathology reports, and then these pre-trained embeddings are used as input for a multi-task convolutional neural network (MT-CNN) \cite{Alawad2018BHI}. This approach uses a semantically informed method but disregards the valuable information provided by domain knowledge.

The integration of domain knowledge happens as part of the initial training of the word embeddings. Specifically, lexico-semantic information originally captured by the word embeddings techniques can be further enhanced with relational information extracted from the domain knowledge base. This domain knowledge is commonly represented in the form of ontologies such as unified medical language system (UMLS). Previous work has used domain-specific information to complement the biomedical corpus in multiple ways, such as fine-tune pre-trained embeddings, UMLS Concept User Identifier Embedding, and Code Embedding \cite{kalyan2020secnlp}. In this paper, for the integration, we enrich the pre-trained embeddings using UMLS as the domain knowledge. The enriching of word embeddings is done by updating the distributional representative space based on the semantic similarity between two words established by the original word embedding technique and the biomedical taxonomy from UMLS \cite{Alawad2018BigData}. 
The goal for the enriched word embeddings is for conceptually similar words, such has ``Renal Failure" and ``Kidney Failure", to be closer to one another even if they have low co-occurrence in the pretraining corpora. This enriching method helps reduce issues caused by linguistic variability that is prevalent across different pathologists. This enriched embedding is used as the input for our MT-CNN model. Based on our previous work, the MT-CNN was shown to be an efficient model for information extraction tasks in the cancer domain \cite{Alawad2018BHI}. 

In this paper, we propose an MT-CNN built using UMLS enriched word embeddings. The MT-CNN is designed to perform information extraction from cancer pathology reports on six cancer key characteristics -- site, subsite, laterality, behavior, histology and grade. These information extraction tasks were performed on ~900k cancer pathology reports obtained through the SEER Program\footnote{https://seer.cancer.gov/}. The performance of the proposed domain-enriched embeddings are evaluated by comparing F1 scores between the MT-CNN using word embedding with and without enrichment.

\section{Materials and Methods}
\label{methods}
\subsection{Cancer Pathology Reports}
\label{path}
Our dataset consists of cancer pathology reports obtained from the Louisiana Tumor Registry (LTR), Kentucky Cancer Registry (KCR), Utah Cancer Registry (UCR), and New Jersey State Cancer Registry (NJSCR) of the SEER Program\footnote{NJSCR is no longer in the SEER Program, but is included in the current data release.}. The study is executed in accordance to the institutional review board protocol DOE000152. Each pathology report in our dataset is associated with a unique tumor ID; the same tumor ID may be associated with one or more pathology reports. For each tumor ID, Certified Tumor Registrars (CTRs) manually assigned ground truth labels for key data elements -- cancer site, histology, grade, etc. -- based off all data available for that tumor ID according to the SEER program coding and staging manual\footnote{https://seer.cancer.gov/tools/codingmanuals/index.html}. In this paper, site (70 classes), subsite (324 classes), laterality (7 classes), behavior (4 classes), histology (572 classes), and grade (9 classes) are defined as the target classification labels for each cancer pathology report as these are fundamental information extraction tasks for cancer phenotyping. Documents generated within 10 days between the date of pathological diagnosis and either path specimen collection date or the surgery date are identified as relevant to the specific case ID. 
We split the dataset based on the pathology report specimen collection date into train, validation and test sets. Specifically, reports collected prior to 2017 are used for train and validation with 80:20 ratio, while the rest of the reports are used for testing.
The resulting train, validation, and test datasets consist of 598,228, 149,315, and 131,322 pathology reports respectively, yielding a total of 878,865 documents for our experiment.

For each pathology report, we extract the text content and apply standard preprocessing steps to tokenize it. First, we remove XML tags as well as identification tags. In addition, we set all alphabetical characters to lowercase. Then, we create a vocabulary list including all words that have pathology report document frequency of at least five. 
Since pathology reports have different lengths, we accommodated this by specifying a fixed length of $3,000$ words for all reports.

\subsection{UMLS Metathesaurus Vocabularies}
The UMLS is a repository of files and software that integrate various biomedical vocabularies and standards developed by the National Library of Medicine (NLM)~\cite{umls-tutorial}. The motivation behind the development of the UMLS is to help computer programs to understand the semantics of biomedicine and health~\cite{lindberg1993unified}. UMLS provides three knowledge sources. First, the Metathesaurus that contains numerous vocabulary terms and codes. Second, the Semantic Network provides semantic types and relations. Third, it contains the Specialist Lexicon for natural language processing~\cite{umls-tutorial}. 
In this paper, we use three vocabulary sources to build our knowledge graph: i) Systematized Nomenclature of Medicine-Clinical Terms (SNOMED CT US Edition), which provides major general terminology for the electronic health record (EHR) and has hierarchies made from specific conceptual meanings and formal logic-based definitions~\cite{snomed-synopsis}. ii) National Cancer Institute (NCI) Thesaurus,  which is a product of NCI Enterprise Vocabulary Services (EVS) and includes the vocabularies about clinical cancer care, translational, and basic research. It consists of public information on cancer together with administrative activity vocabularies~\cite{nci-synopsis}. iii) International Classification of Diseases, Tenth Revision (ICD10)~\cite{umls-vocab-doc}, which is introduced and maintained by the World Health Organization (WHO). It is made up of titles and codes for causes of death, inclusion and exclusion terms for cause-of-death titles, an alphabetical index to diseases and nature of injury, external causes of injury, table of drugs and chemicals, and description, guidelines, and coding rules~\cite{icd10-synopsis}. 

To build our graph, we download the latest version of the UMLS release (2018AA) from~\cite{umls-release-files} for local installation. It contains 3,665,926 concepts from 154 sources~\cite{umls-stat-2018}. We create MySQL database load scripts by using the UMLS installation tool. 
Then, we develop a Python software tool that reads words from the vocabulary list (mentioned in Section~\ref{path}), connects them to MySQL database, and fetches the similar UMLS concept names by using the SQL LIKE operator. The software collects all similar UMLS concept names from three vocabularies mentioned earlier (SNOMED CT, NCI, and ICD-10). Concept names are ordered by the UMLS concept unique identifiers (CUIs). We observe that in the UMLS database sometimes one CUI represents multiple similar concept names. In that situation, we use the first concept name.

\subsection{Enrich Word Embeddings with Domain Knowledge}
\label{subsection:retrofitting}

The enhanced word embeddings used in this paper are pre-trained in two phases. In phase one, the word embeddings are first trained using traditional distributional similarity approaches such as word2vec without incorporating the domain knowledge from UMLS knowledge graph. The outcome is pre-trained word vectors corresponding to the set of vocabularies $V = \{v_1,v_2 \dots ,v_n\}$ from the cancer pathology reports. In the second phase, the retrofitting algorithm, presented in~\cite{Faruqui2015}, is used to refine the pre-trained word embeddings resulting from the first phase. Specifically, a undirected graph is created from the vocabulary words $V$, the vertices of the graph, and the UMLS graph that finds the relationships between clinical  concepts in $V$ and defines the edges $E$ of the graph. Then, the retrofitted word embeddings $\hat W$are initialized to be equal to the word embeddings $W$ resulting from phase one. The objective is to update the word vectors to be close to their counterparts in $W$ and close to the adjacent vertices in the graph. The distance that needs to be minimized to ascertain these objectives can be defined as follows:
\begin{equation}
\label{eq1}
\Psi = \sum_{i=1}^n \Bigg\lbrack \alpha_i ||\hat{w}-w||^2+\sum_{(i,j)\in E} \beta_{ij}||\hat{w_i}-\hat{w_j}||^2\Bigg\rbrack
\end{equation}
Where $n$ is the vocabulary list size, $w_i$ is the feature representation of word $v_i$, $\hat w_i$ is the retrofitted feature representation of word $v_i$, $w_j$ is the feature representation of the adjacent words in the graph to the word $v_i$, where $(i,j) \in E$. $\hat W$ words vectors are updated by taking the first derivative of equation~\ref{eq1} with respect to $\hat w_i$ as follows:
\begin{equation}
\hat{w_i} = \frac{\sum_{j:(i,j)\in E} \beta_{ij}\hat{w_j} + \alpha_i w_i}{\sum_{j:(i,j)\in E} \beta_{ij} + \alpha_i}
\end{equation}
This method can be used to update word embeddings produced from any model. Therefore, we can use any existing method to pre-train word embeddings and then apply our retrofitting algorithm to enrich the pre-trained word embedding vectors with domain knowledge from a relevant knowledge graph.

\subsection{Deep Learning Model}

The deep learning (DL) model used in this paper is a previously developed MT-CNN model~\cite{Alawad2018BHI}. The model consists of three parallel 1D convolution layers that run across the document matrix, i.e. the word embeddings for a given document. These convolution layers have 300 filters each and use window sizes of 3, 4, and 5 consecutive words to capture features with variable context lengths. The outputs from the convolution layers are concatenated and fed into a max-pooling layer that filters out the most important features from the convolution layers. Finally, the output from the max-pooling layer is fed into six separate soft-max fully connected layers -- one for each cancer characteristic. The Baseline MT-CNN model uses pre-trained word embeddings generated by word2vec. The embeddings are trained on our entire corpus of pathology reports. Our enhanced MT-CNN uses enriched word embeddings -- these consist of the same word2vec embeddings that are enhanced with external knowledge from medical terminology ontologies using our retrofitting algorithm.

\begin{table*}[htbp]
\renewcommand{\arraystretch}{1.0}
\setlength{\tabcolsep}{12pt}
\caption{Performance comparison of classification models. The non-bold numbers under each bold number are 95\% confidence intervals}
\begin{center}
\begin{tabular}{|c|c|c|c|c|c|c|c|c|}
\hline
\rowcolor[HTML]{EFEFEF}
\diaghead{\theadfont Diag ColumnmnHead II}%
  {\textbf{Task}}{\textbf{DL Model} \\ \textbf{Type}} &\multicolumn{4}{c|}{\textbf{Baseline MT-CNN}} &\multicolumn{4}{c|}{\textbf{MT-CNN with UMLS Knowledge Graph}}\\
\hline
\rowcolor[HTML]{EFEFEF}
 & \multicolumn{2}{c|}{\textbf{micro-F1}}& \multicolumn{2}{c|}{\textbf{macro-F1}} & \multicolumn{2}{c|}{\textbf{micro-F1}} & \multicolumn{2}{c|}{\textbf{macro-F1}} \\
\hline
\multirow{2}{*}{\textbf{Site}} & \multicolumn{2}{c|}{\textbf{0.916}}& \multicolumn{2}{c|}{\textbf{0.593}} & \multicolumn{2}{c|}{\textbf{0.934}} & \multicolumn{2}{c|}{\textbf{0.703}} \\
\cline{2-9} 
 & 0.914 &	0.917 &	0.585 &	0.604 &	0.932 &	0.935 &	0.689 &	0.713 \\
\hline
\multirow{2}{*}{\textbf{Subsite}} & \multicolumn{2}{c|}{\textbf{0.596}}& \multicolumn{2}{c|}{\textbf{0.228}} & \multicolumn{2}{c|}{\textbf{0.681}} & \multicolumn{2}{c|}{\textbf{0.334}} \\
\cline{2-9} 
 & 0.594 &	0.599 &	0.228 &	0.239 &	0.678 &	0.683 &	0.332 &	0.348 \\
\hline
\multirow{2}{*}{\textbf{Laterality}} & \multicolumn{2}{c|}{\textbf{0.890}}& \multicolumn{2}{c|}{\textbf{0.487}} & \multicolumn{2}{c|}{\textbf{0.918}} & \multicolumn{2}{c|}{\textbf{0.536}} \\
\cline{2-9} 
 & 0.888 &	0.892 &	0.480 &	0.495 &	0.917 &	0.920 &	0.529 &	0.543 \\
\hline
\multirow{2}{*}{\textbf{Histology}} & \multicolumn{2}{c|}{\textbf{0.756}}& \multicolumn{2}{c|}{\textbf{0.235}} & \multicolumn{2}{c|}{\textbf{0.787}} & \multicolumn{2}{c|}{\textbf{0.375}} \\
\cline{2-9} 
 & 0.754 &	0.759 &	0.238 &	0.250 &	0.785  & 0.790 &	0.374 &	0.392 \\
\hline
\multirow{2}{*}{\textbf{Behavior}} & \multicolumn{2}{c|}{\textbf{0.959}}& \multicolumn{2}{c|}{\textbf{0.780}} & \multicolumn{2}{c|}{\textbf{0.974}} & \multicolumn{2}{c|}{\textbf{0.897}} \\
\cline{2-9} 
 & 0.957 &	0.960 &	0.762 &	0.798 &	0.973 &	0.975 &	0.884 &	0.908 \\
\hline
\multirow{2}{*}{\textbf{Grade}} & \multicolumn{2}{c|}{\textbf{0.716}}& \multicolumn{2}{c|}{\textbf{0.559}} & \multicolumn{2}{c|}{\textbf{0.776}} & \multicolumn{2}{c|}{\textbf{0.682}} \\
\cline{2-9} 
 & 0.713 &	0.718 &	0.555 &	0.562 &	0.774 &	0.778 &	0.661 &	0.699 \\
\hline
\end{tabular}
\label{tab:results}
\end{center}
\end{table*}

\section{Performance Evaluation and Experimental Results}
\label{results}

In our experiments, we compare our proposed MT-CNN model that uses enhanced word embeddings retrofitted from the UMLS knowledge graph against the baseline MT-CNN model that uses pre-trained word2vec embeddings. 
We use the pre-trained embeddings without any adjustments to see the advantage of our proposed approach. For both the models used in this paper, word embeddings are fine-tuned via back propagation along with the other model parameters.

To evaluate the performance accuracy of the DL models, we use the standard NLP metrics of micro- and
macro-F1 scores. The micro-F1 score has class representation
roughly proportional to their test set representation, whereas the
macro-F1 score is averaged by class without
weighing by class prevalence. Thus, it better shows the impact of different DL models on the low prevalence classes.

\subsection{Experimental Results}

In this work, we tested our proposed method on automated information extraction from cancer pathology reports for six cancer key characteristics -- site, subsite, laterality, behavior, histology, and grade. We developed an MT-CNN model to extract all six tasks simultaneously. We built a graph based on the vocabulary list from the cancer pathology reports and the UMLS knowledge graph that combines three vocabulary resources -- SNOMED, NCI, and ICD10. The size of the graph  is 59,604 nodes and 239,665 edges. For all the experiments, we included 95\% confidence intervals for each performance metric derived using bootstrapping~\cite{efron1994}. 

The performance results on all six tasks are summarized in Table~\ref{tab:results}. The results show that MT-CNN model with word embeddings enriched by UMLS knowledge graph consistently outperform the the baseline MT-CNN model that uses pre-trained word embeddings using word2vec. The average performance of the baseline DL model micro- and macro-averaged F1 scores are 0.805 and 0.480, respectively. While the proposed approach achieves  micro- and macro-F1 scores of 0.845 and 0.588; that is an improvement of 4.97\% and 22.5\% for micro- and macro-F1 scores, respectively.

\begin{figure}[h]
\centering
  
  \includegraphics[width=\linewidth]{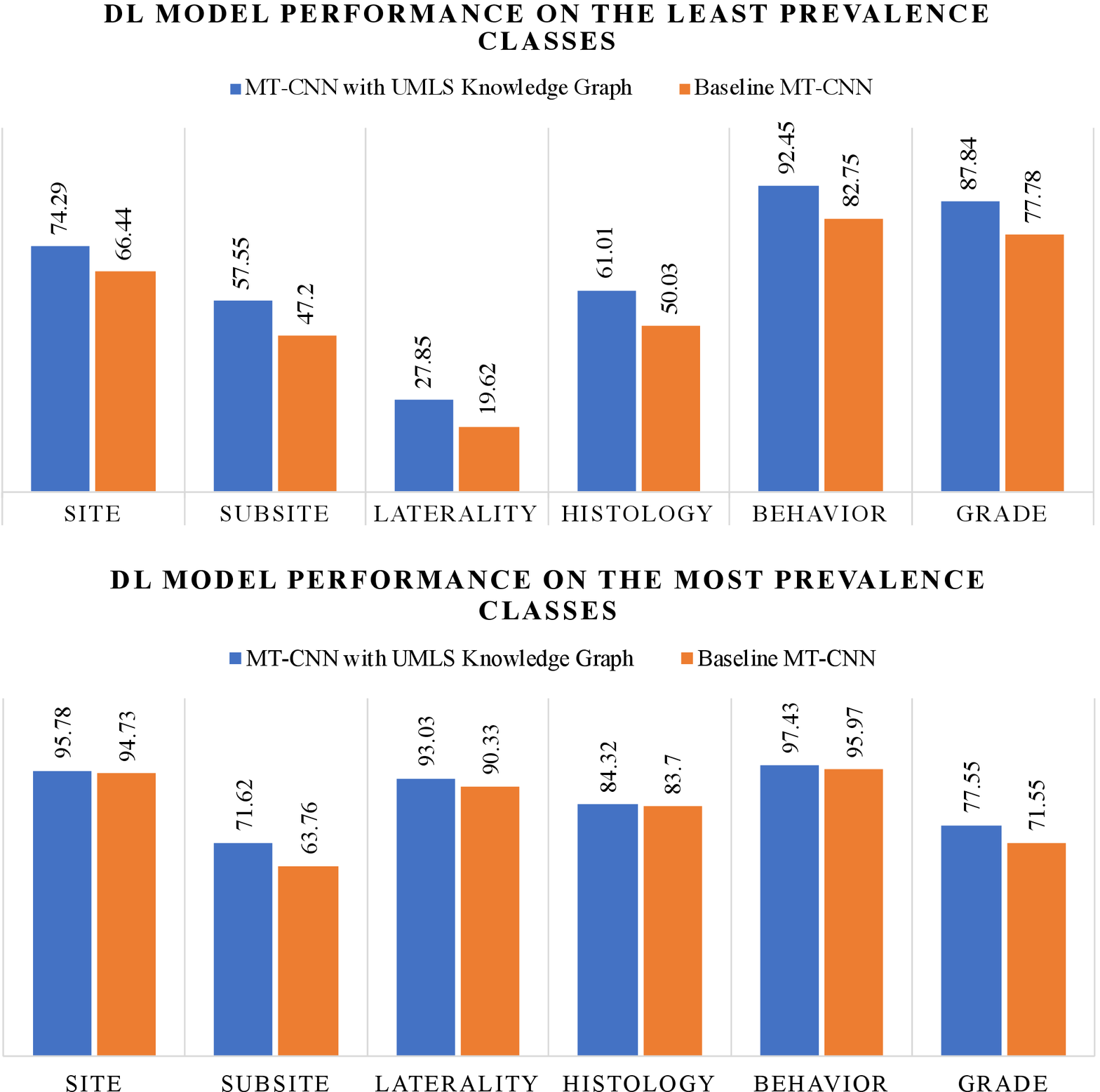}
  \caption{DL model performance on most and least prevalence classes.}
\label{fig:imbalance}
\end{figure}

Class imbalance is one of the challenges in real-world datasets, especially in clinical domains. Our cancer pathology report dataset has  extreme  variability  in  class  prevalence. This is a common challenge in cancer registries because some  cancer  types  are  highly  prevalent  (e.g.,  breast,  lung,  prostate) while others are very rare (e.g.,  esophagus,  gum,  sinuses).  To study in more detail the impact of class prevalence on the classification accuracy, we show in Figure~\ref{fig:imbalance} the average accuracy of the two models on the most prevalence and the least prevalence classes across all six tasks. As expected, both the models perform the best on the more prevalence classes. However, our proposed approach outperform the baseline MT-CNN on the low prevalence classes. These results highlight the advantage of incorporating domain knowledge with deep learning to boost the classification performance on the low prevalence classes by enriching the text representation with external knowledge.
 
Finally, we study the performance of our proposed MT-CNN and the baseline MT-CNN models on predicting cancer phenotype, i.e. correctly classifying multiple cancer characteristics for a given document. When considering all six cancer characteristics in this six, the model with retrofitted embeddings correctly phenotypes 41.66\% of the pathology reports, while the baseline model correctly phenotypes only 31.47\% of the pathology reports. When considering the site, laterality, behavior, and histology characteristics, the proposed model correctly phenotypes 68.89\% of the pathology reports, while the baseline model correctly phenotypes only 62.88\% of the pathology reports.

\section{Conclusion}
\label{conclusion}
This paper proposes the use of external domain knowledge to enrich the word embeddings of deep learning models used for NLP tasks. The model is used for computational cancer phenotyping. Compared to the baseline MT-CNN, our proposed approach achieves a statistically significant higher accuracy across all tasks. This suggests that information extraction tasks, which are fundamental for cancer surveillance programs, may benefit from enriching the data with external knowledge. Using UMLS knowledge graph to enrich word representations helps in tackling the class imbalance issue by significantly improving the macro-F1 score.

\section*{Acknowledgment}
This work has been supported in part by the Joint Design of Advanced Computing Solutions for Cancer (JDACS4C) program established by the U.S. Department of Energy (DOE) and the National Cancer Institute (NCI) of the National Institutes of Health. This work was performed under the auspices of the U.S. Department of Energy by Argonne National Laboratory under Contract DE-AC02-06-CH11357, Lawrence Livermore National Laboratory under Contract DEAC52-07NA27344, Los Alamos National Laboratory under Contract DE-AC5206NA25396, and Oak Ridge National Laboratory under Contract DE-AC05-00OR22725. 

This work has also been supported by National Cancer Institute under Contract No. HHSN261201800013I/HHSN26100001 and NCI Cancer Center Support Grant (P30CA177558).

\bibliographystyle{ieeetr} 

\bibliography{references}

\balance

\end{document}